# Semantic BIM enrichment for firefighting assets: Fire-ART dataset and panoramic image-based 3D reconstruction

Ya Wen [a], Yutong Qiao [b], Chi Chiu Lam [b], Ioannis Brilakis [a], Sanghoon Lee [c], and Mun On Wong [b, *]

[a] Department of Engineering, University of Cambridge, Cambridge, United Kingdom
[b] Department of Civil and Environmental Engineering, University of Macau, Macau SAR, China
[c] Department of Architectural Engineering, University of Seoul, Seoul, South Korea
Email: yw710@cam.ac.uk, yutong.qiao@connect.um.edu.mo, fstccl@um.edu.mo, ib340@cam.ac.uk, sanghoon.lee@uos.ac.kr, mowong@um.edu.mo
* Corresponding author.

## Abstract

Inventory management of firefighting assets is crucial for emergency preparedness, risk assessment, and on-site fire response. However, conventional methods are inefficient due to limited capabilities in automated asset recognition and reconstruction. To address the challenge, this research introduces the Fire-ART dataset and develops a panoramic image-based reconstruction approach for semantic enrichment of firefighting assets into BIM models. The Fire-ART dataset covers 15 fundamental assets, comprising 2,626 images and 6,627 instances, making it an extensive and publicly accessible dataset for asset recognition. In addition, the reconstruction approach integrates modified cube-map conversion and radius-based spherical camera projection to enhance recognition and localization accuracy. Through validations with two real-world case studies, the proposed approach achieves F1-scores of 73% and 88% and localization errors of 0.620 and 0.428 meters, respectively. The Fire-ART dataset and the reconstruction approach offer valuable resources and robust technical solutions to enhance the accurate digital management of fire safety equipment.

## 1. Introduction

Indoor fire emergency is one of the most critical incidents that cause fatalities, injuries, and property loss. In 2022, fire incidents in the United States resulted in 4,445 deaths and 13,250 injuries (U.S. Fire Administration, 2024a). Furthermore, an analysis of data from 2014 to 2023 reveals a 45.5% increase in fire incidents and a 63.5% rise in financial losses across residential and non-residential buildings, primarily due to unintentional or careless fires (U.S. Fire Administration, 2024b, 2024c). To mitigate fire threats, effective fire safety management is vital to enable early identification of any potential risks and facilitate timely information management for on-site emergency responses (Wong and Lee, 2022). In fire safety management, accurate and comprehensive information of firefighting assets are of most importance to enhance the stakeholders' awareness of available resources for informed decision-making. These data support a diverse array of applications, including fire risk assessments (Qiao et al., 2024a, 2024b), code compliance verification (Fitkau and Hartmann, 2024), fire dynamic simulations (Chen et al., 2018), and real-time emergency response (Ma and Wu, 2020). For instance, indicators derived from the quantity, type, and spatial distribution of firefighting assets provide objective metrics for assessing risk levels in both modern and historical structures (Qiao et al., 2024a, 2024b). Detailed information on the locations and operational statuses of fire service installations further enhances evacuation strategies and fire suppression efforts to benefit both building occupants and first responders (Wong et al., 2022; Wong and Lee, 2023).

In practice, firefighting assets include various equipment such as alarms, sprinklers, extinguishers, and hose reels, spreading across different locations in indoor environments. Information about these critical components is traditionally recorded in two-dimensional (2D) drawings. This conventional approach limits the ability to interpret data digitally and thus hinders the application of data-driven analysis for fire safety management (Wang et al., 2015). In recent years, building information modeling (BIM) and digital twin (DT) technologies have emerged as potential solutions for digital transformation. Many efforts have been made to develop BIM- or DT-driven approaches for fire safety design and management (Boje et al., 2020). In particular, for newly constructed buildings, the building information can be modeled with comprehensive data about the firefighting assets simultaneously to facilitate advanced planning, simulations, and code compliance checking. However, the majority of existing buildings were constructed before the advent of these digital technologies, meaning that they usually lack BIM models or have only basic architectural models. As a result, essential information about firefighting assets is often missing (Li et al., 2014). While conventional BIM authoring tools have functionalities to enrich BIM with firefighting asset information, such a process usually involves intensive manual efforts and is time-consuming (Wang et al., 2015).

To overcome this inefficiency, researchers have explored automated techniques to gather firefighting asset data from either 2D drawings or real-world settings. For example, Bortoluzzi et al. (2019) utilized 2D floor plans and elevation drawings to create semantically-rich BIM models for facility operation and management. Schönfelder et al. (2024) introduced an automated method to extract fire safety equipment data from 2D escape plans by accurately identifying asset symbol positions and translating them into spatial coordinates within BIM models. Although such drawing-based methods are promising, the enrichment results may be affected by outdated documentation or building renovative activities and, therefore, they cannot reflect the latest situation of fire service installations in real-world environments.

In comparison, acquiring data directly from physical environments provides a distinct benefit to yield up-to-date and real-time insights into asset conditions (Zhou et al., 2022). Increasing attention has been paid to exploring reality capture technologies to reconstruct different indoor building assets (Meyer et al., 2023; Yang et al., 2021). For instance, Di and Gong (2024) utilized point cloud data from static laser scanner to extract safety-related objects like security camera, ID reader, door exit push, and fire alarm for inventory management in school buildings. Pan et al. (2022) fused laser scanning and photogrammetry methods to capture, recognize, and map small objects in buildings, such as light switch, trash bin, escape sign, and socket for geometric DT enrichment. Nevertheless, the scope of prior studies rarely centers on fire safety, leaving the key firefighting components understudied.

More precisely, two limitations exist in the current reality capture-based methods for firefighting assets recognition and reconstruction. First, there is a scarcity of datasets designed for recognizing firefighting assets. While many existing datasets, such as COCO (Lin et al., 2014) and ImageNet (Deng et al., 2009) cover general objects in daily scenes and indoor environments, they contain only a few firefighting assets separately, which make them unsuitable for comprehensive fire safety applications. Instead of general-purpose datasets, most existing studies have dominantly utilized the FireNet dataset (Boehm et al., 2019), which contains 1,452 images spanning 8 basic categories (Bayer and Aziz, 2022). While this dataset already provides a valuable resource, several critical items, such as hose reels, sprinklers, and emergency lights, are not included, leaving room for further improvement.

Second, a cost-effective and highly accessible solution for firefighting asset reconstruction and BIM enrichment remains largely uninvestigated. Current research frequently relies on precise yet expensive technologies, such as light detection and ranging (LiDAR) or laser scanners, to capture indoor scenes (Adán et al., 2018). While these methods provide high accuracy, their high costs and specialized equipment can limit their widespread adoption, especially for smaller-scale projects or budget-conscious applications. By contrast, photogrammetry using cameras offers a promising alternative. Through structure from motion and multi-view stereo techniques, 3D point clouds or meshes of the indoor scenes can be generated using a series of overlapped images captured from multiple angles. In particular, 360-degree cameras are advantageous to capture full spatial scenes for accurately collecting firefighting asset data. Despite its potential benefits, the sole utilization of panoramic photogrammetry for firefighting asset reconstruction has received very limited attention. Specifically, the mechanism of converting firefighting assets in video frames into information-enriched BIM models is still unclear and deserves further investigation (Wei and Akinci, 2022).

To address the identified shortcomings, this research proposes a novel dataset and a 3D semantic reconstruction approach for enriching firefighting assets into BIM models. The firefighting asset recognition dataset, Fire-ART, is created in this research, comprising 15 common categories of fire equipment, including fire extinguisher, fire exit sign, fire door sign, fire alarm, emergency light, smoke detector, fire hose reel, piping system, sprinkler, fire call point, emergency door release, fire blanket, fire equipment sign, firefighting lift switch, and hidden fire equipment. The dataset includes 2,626 images and 6,627 instances in total. Specifically, the dataset offers unique advantages by incorporating more fundamental asset types, encompassing a larger number of images and associated annotations, and covering images from multiple sources across various geographical locations and space types.

In addition, this research establishes a novel approach that leverages the strengths of panoramic images for photogrammetric reconstruction and overcomes image distortion for accurate 3D semantic segmentation of firefighting assets. The proposed approach, with rich visual coverage, is effective in reconstructing indoor scenes and documenting the semantic classes and the corresponding locations of the firefighting assets. Through two real-world case studies in the UK and Macau SAR, the approach is validated feasible and effective, achieving precision rates of 82% and 94% as well as recall rates of 67% and 84%, respectively, with localization accuracy of 0.620 meters and 0.428 meters. By integrating the Fire-ART dataset and the proposed approach, this research offers a pioneering solution to enhance timely and accurate digital management of fire safety equipment.

The structure of the paper is as follows. Section 2 critically reviews the prior research on firefighting asset recognition and the existing datasets of firefighting asset images, offering a detailed synthesis of their strengths and limitations. It further explores methodologies for BIM enrichment using firefighting-related data. Section 3 elucidates the creation of the Fire-ART dataset and introduces the proposed panoramic image-based reconstruction approach for BIM enrichment with firefighting asset information. Section 4 presents the implementation of two case studies and illustrates the practical applications of the dataset and the proposed approach with in-depth analysis of its accuracy and performance. Sections 5 and 6 examine key challenges, broader implications, inherent limitations, and prospective avenues for future research.

## 2. Literature review

### 2.1 Firefighting asset recognition

Recognizing various firefighting assets is vital for effective fire safety management. Accurate information on the quantity and spatial distribution of these assets supports a range of safety-related applications, such as emergency preparedness, fire risk assessment, and informed fire response. With the development of computer vision technologies, supervised learning-based approaches have become increasingly prevalent in facilitating accurate object recognition within images. Specifically, the effectiveness of these methods heavily depends on the scale and quality of annotated datasets. Many general-purpose datasets, such as ImageNet (Deng et al., 2009) and COCO (Lin et al., 2014) exist to support various visual recognition tasks for interior and exterior scenes. For instance, ImageNet is a large-scale image database for visual object classification, from which several fire-related classes such as fire engine and fireboat are included. Similarly, the COCO dataset is designed for the detection of 80 common object categories encountered in daily life, including fire hydrant systems. However, since these general-purpose datasets were not designed for firefighting asset recognition, they only have limited relevant data, thereby restricting their direct utilization to this domain.

To address this challenge, a prior study proposed the FireNet dataset (Boehm et al., 2019) specialized for recognizing firefighting assets. The FireNet dataset contains images of daily scenes captured in the United Kingdom (UK), covering eight essential fire asset classes, i.e., fire extinguisher, call point, protective blanket, escape route sign, fire equipment sign, visual alarm, fire alarm sounder, and smoke detector. Overall, it comprises over 1,400 images and around 2,000 annotated instances, forming a valuable resource for building safety documentation (Bayer and Aziz, 2022) and equipment inspection (Aziz et al., 2024; Heinbach et al., 2023). Despite its strengths, the dataset still has two limitations. First, its images exclusively originate from UK environments, which restricts its applicability in different geographical regions due to variations in equipment and sign appearance, design standards, and language differences. Second, although FireNet covers eight critical firefighting assets, certain essential types (e.g., hose reel and sprinkler) remain absent, suggesting a need for expansion and further data collection to enhance dataset coverage.

In addition to the FireNet dataset, several existing studies have attempted to develop custom datasets for safety-related object detection. For instance, Corneli et al. (2020) created a custom dataset to facilitate detection of fire extinguishers and emergency signages to streamline the component inventory process. Bayer and Aziz (2022) combined FireNet with a self-created image dataset to enable the detection of fire extinguishers, fire call points, smoke detectors, and fire safety blankets to support fire safety management. Pan et al. (2022) collected images of interior office scenes in the Technical University of Munich and created a building small object detection dataset, which includes emergency switches, smoke alarms, fire extinguishers, escape route signs, pipes, and door signs. Similarly, Di and Gong (2024) developed a dataset for safety inventory management by capturing indoor scene images within university buildings at the United States and labeling ten object classes, while seven of which specifically pertain to fire safety components. Lin et al. (2025) developed an RGB-D dataset using images captured in Hong Kong to conduct 3D spatial detection of eight firefighting asset types. Table 1 summarizes the characteristics of the existing open and custom datasets for firefighting asset recognition. While these studies have demonstrated promising results, the proposed custom datasets commonly share several critical limitations. First, they are usually captured within limited scenes and buildings in specific regions, which constrain their generalizability to other geographic areas. Second, although most datasets cover a certain range of asset classes, several critical firefighting assets remain lacking. Last, these datasets are not directly accessible to the public and thus restrict their utilizations for further research and practical applications.

In response to the limitations of region-specific coverage, insufficient asset types, and inaccessibility to the datasets, it is necessary to integrate and expand existing resources by including more asset types with additional labeled data captured from diverse countries and regions.

Table 1. Datasets related to firefighting asset recognition

| Dataset | Data type | Recognition type | No. of class | No. of images | No. of instances | Relevant firefighting asset classes | Availability |
|---|---|---|---|---|---|---|---|
| ImageNet-1k (Deng et al., 2009) | RGB images | Object classification | 1,000 | 1,431,167 | N.A. | Fire engine, fireguard, and fireboat | Public |
| COCO (Lin et al., 2014) | RGB images | Instance segmentation | 80 | 123,287 | 886,284 | Fire hydrant | Public |
| FireNet (Boehm et al., 2019) | RGB images | Semantic segmentation and object detection | 8 | 1,452 | 2,154 | Fire extinguisher, call point, protective blanket, escape route sign, fire equipment sign, visual alarm, fire alarm sounder, and smoke detector | Public |
| Corneli et al. (2020) | RGB images | Object detection | 4 | 1,581 | N.A. | Extinguisher and emergency sign (emergency sign door, emergency sign man, and emergency sign) | Not public |
| Bayer and Aziz (2022) | RGB images | Object detection | 4 | 841 | N.A. | Fire extinguisher, emergency call point, smoke detector, and fire safety blanket | Not public |
| Pan et al. (2022) | RGB images | Instance segmentation | 12 | N/A | 1241 | Emergency switch, smoke alarm, fire extinguisher, escape sign, pipes, and door sign | Not public |
| Di and Gong (2024) | RGB panoramic images | Object detection | 10 | 2089 (training) | 3166 (training) | Smoke alarm, emergency light, exit sign, fire alarm, fire alarm switch, fire extinguisher, and exit push | Not public |
| Lin et al. (2025) | RGB-D images | Object detection | 8 | 2,793 | 6,450 | Alarm bell, alarm button, fire extinguisher, escape sign, emergency shower, hose reel, warning light, and sand bucket | Not public |
| Fire-ART (Ours) | RGB images | Instance segmentation | 15 | 2,626 | 6,627 | Fire extinguisher, fire exit sign, fire door sign, fire alarm, emergency light, smoke detector, fire hose reel, piping system, sprinkler, fire call point, emergency door release, fire blanket, fire equipment sign, firefighting lift switch, and hidden fire equipment | Public |

## 2.2 BIM enrichment of firefighting assets

In current practice, most as-designed and as-is BIM models of existing buildings lack detailed information on firefighting assets due to the considerable efforts required for manual model compilation. Semantic BIM enrichment addresses this gap by integrating meaningful semantic data into the digital representation of built assets automatically. This process involves constructing or updating BIM models to reflect the current physical conditions of a building so as to enhance the accuracy and utility of the model for facility management and safety applications.

To enable BIM reconstruction and enrichment, existing research has developed different approaches using various data sources, including drawings, laser scanning point clouds, images, and videos (Bloch and Sacks, 2018). Specifically, there are two mainstream methods, i.e., drawing-based and reality capture-based approaches. The drawing-based approach leverages design and handover drawings to provide accurate representations of building components, serving as critical inputs for generating or updating 3D BIM models (Cho and Liu, 2017; Yin et al., 2020; Zhao et al., 2021). For instance, Cho and Liu (2017) developed a method to reconstruct mechanical systems in 3D BIM models by detecting component patterns, texts, and annotations based on predefined mechanical system graphs. Similarly, Yin et al. (2020) extracted location data from floor plan drawings and height information from elevation drawings to construct BIM models. Additionally, Zhao et al. (2021) employed a faster region-based convolutional neural network (Faster R-CNN) to detect objects in architectural drawings, complemented by optical character recognition (OCR) to extract geometric location information of these objects. Bortoluzzi et al. (2019) developed algorithms to recognize building exterior boundaries, individual room boundaries, and room name tags for automatic BIM modeling of buildings from 2D floor plans and elevation drawings to empower facility management. Regarding semantic firefighting information, Schönfelder et al. (2024) developed an automated workflow to interpret fire safety equipment information from 2D escape plans and then modeled fire safety objects manually in the BIM model.

For the reality capture-based approach, existing research usually leverages LiDAR, laser scanning, or photogrammetry techniques to scan the physical environments and then recognize interested building objects for semantic enrichment (Huan et al., 2022; Hübner et al., 2021). For instance, Adán et al. (2018) combined laser point clouds and images to recognize secondary building objects such as electrical panels, sockets, and switches to enrich the details of reconstructed BIM models. Similarly, Pan et al. (2022) employed laser scanning and photogrammetry techniques to detect 12 small-scale building objects like light switches, lights, speakers, sockets, and elevator buttons within interior scenes for enhancing the level of developments of geometric digital twins. Specifically, several studies focused on the enrichment of safety-related elements. Di and Gong (2024) leveraged stationary laser scanners with 360 cameras to perform object detection of safety elements (e.g., CCTVs, exit signs, exit pushes, and ID locks) for inventory management of school buildings. Lin et al. (2025) integrated the depth maps with RGB images to develop a modified YOLO model for enhanced fire safety equipment inspection. Ding et al. (2025) proposed a novel fuel load dataset and leveraged image segmentation and manual room localization to assess the spatial distribution of indoor fuel loads and potential fire spread hazards. Table 2 summarizes the existing efforts of both drawing- and reality capture-based methods for building semantic enrichment.

Table 2. Existing studies related to semantic BIM enrichment

| Research | Data source | Scope | Semantic enrichment mechanism |
|---|---|---|---|
| Cho and Liu (2017) | 2D mechanical drawings | Mechanical and annotation elements, including duct, duct elbow, duct branch, and pipe | Minimum circuit search method and computer vision to recognize various components |
| Yin et al. (2020) | 2D elevation drawings | Elevation and façade information, including façade orientation, floor levels, coordinates, walls, and façade openings. | Self-developed automatic layer classification method and elevation detection method |
| Zhao et al. (2021) | 2D structural drawings | Structural and annotation elements, including grid head, column, and beam | Faster R-CNN with OCR |
| Bortoluzzi et al. (2019) | 2D floorplans and elevation drawings | Architectural elements, including walls, slabs, openings, and rooms | Self-developed scripts for recognition of building exterior boundaries, individual room boundaries, and room number tags. |
| Schönfelder et al. (2024) | 2D escape plans | Fire safety equipment, including fire extinguisher, manual call point, fire hose reel, firefighting equipment, emergency photo. and fire ladder. | Keypoint R-CNN model for symbol detection |
| Adán et al. (2018) | Laser scanning with images | 10 secondary building components, including electrical panel, socket, switch, fire extinguisher, radiator, fire alarm switch, smoke detector, exit light, extinguisher sign, and fire alarm switch sign | Geometry discontinuity algorithm for detecting protruding objects on walls, as well as pattern classifier for recognition |
| Pan et al. (2022) | Laser scanning with images | 12 small building objects, including light switch, light, speaker, socket, elevator button, emergency switch, smoke alarm, fire extinguisher, escape sign, pipes, door sign, and board. | Detectron2 for image instance segmentation with fusion between image masks and laser point clouds using camera projection as well as OCR for textual enrichment |
| Lin et al. (2025) | RGB-D images | 8 fire safety equipment, including alarm bell, alarm button, fire extinguisher, escape sign, emergency shower, hose reel, warning light, sand bucket | Self-developed YOLO-FSE algorithm for object detection using RGB-D images |
| Ding et al. (2025) | Images | 17 fire load objects, including furniture and electronic elements | YOLOv8-seg for image instance segmentation with manual room selection for rough localization |
| Di and Gong (2024) | Laser scanning with panoramic images | 10 safety-related objects, including door, CCTV, smoke alarm, emergency light, exit sign, fire alarm, fire alarm switch, fire extinguisher, exit push, and ID lock | YOLOv5 for object detection with fusion between image masks and laser point clouds using camera projection |
| Ours | Panoramic images | 15 firefighting asset objects, including Fire extinguisher, fire exit sign, fire door sign, fire alarm, emergency light, smoke detector, fire hose reel, piping system, sprinkler, fire call point, emergency door release, fire blanket, fire equipment sign, firefighting lift switch, and hidden fire equipment | YOLOv8-seg for image instance segmentation with radius-based spherical camera projection and modified cube-map conversion for recognition enhancement |

While existing approaches show promising results in reconstructing building objects and enriching their semantic information in BIM models, several challenges and limitations remain, especially for the application of fire safety management:

(1) In the case of drawing-based approaches, 2D drawings may not accurately capture modifications or additions made to buildings over time. As a result, relying on such drawings for BIM enrichment can lead to outdated or incomplete representations of the physical environment.
(2) Reality capture-based approaches primarily rely on LiDAR and laser scanning technologies, which are recognized for producing highly accurate geometric data. However, their widespread adoption is constrained by the high cost and the need for specialized equipment. In contrast, panoramic photogrammetry with consumer-grade 360-degree cameras offers a low-cost and efficient alternative for capturing complete visual scenes. Despite its potential, research on panoramic photogrammetry as a standalone method for BIM enrichment remains limited, and a systematic exploration of its capabilities, limitations, and practical applications is still lacking.

To address the above limitations, this research proposes an integrated semantic enrichment approach that combines deep learning-based instance segmentation and photogrammetric reconstruction to capture, recognize, reconstruct, and enrich firefighting assets into BIM models. Correspondingly, this research includes two main objectives: (1) to create a new firefighting asset recognition dataset to incorporate more fire safety equipment classes and increase the annotation numbers to facilitate image-based firefighting asset segmentation, and (2) to develop a panoramic image-based approach for enhancing BIM enrichment of firefighting assets through modified cube-map conversion and radius-based spherical camera projection techniques.

## 3. Methodology

This research comprises four critical stages to realize firefighting asset reconstruction and semantic BIM enrichment, i.e., (1) Fire-ART dataset creation, (2) image-based segmentation training, (3) reality capture and 3D semantic reconstruction and (4) BIM enrichment, as shown in Fig. 1.

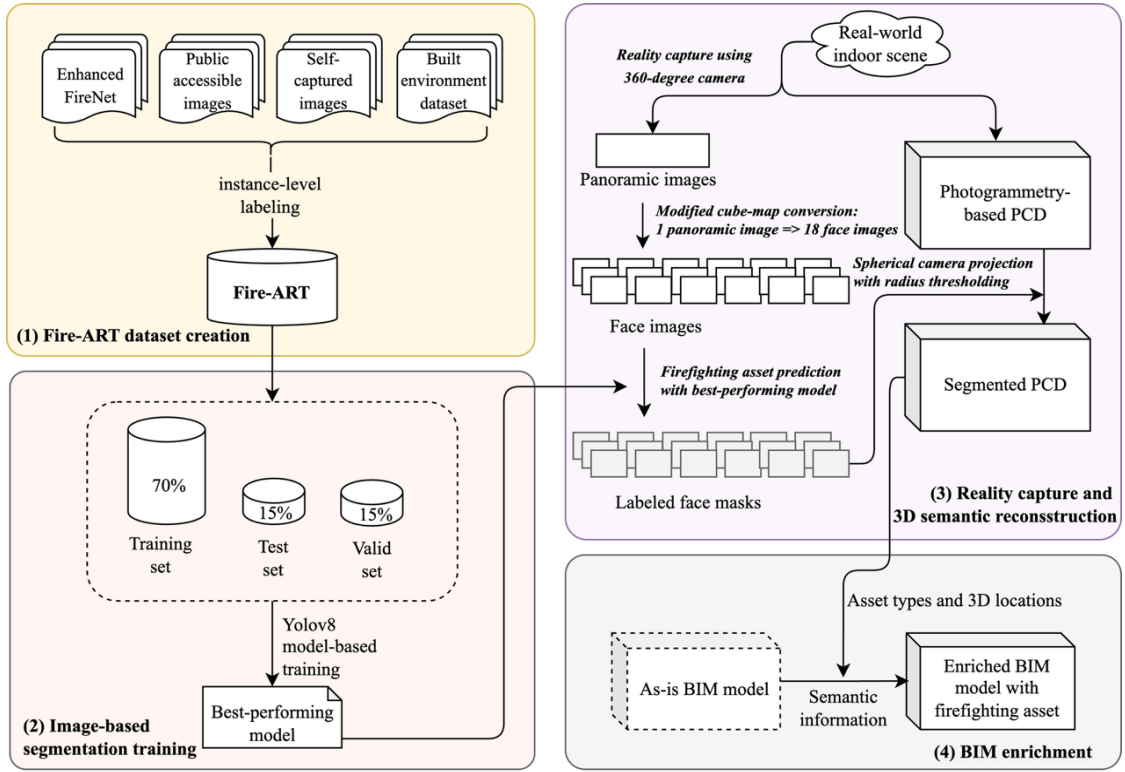

Fig. 1. Methodological workflow of the research

### 3.1 Fire-ART dataset creation

To address the limitations of existing datasets and enhance recognition performance, this research introduces the Fire-ART dataset. Developed by integrating multiple image sources, the Fire-ART dataset aims to ensure diversity, quality, and applicability to real-world firefighting asset recognition scenarios.

3.1.1 Data composition

The Fire-ART dataset comprises four data sources, including (1) enhanced FireNet image set, (2) publicly accessible images, (3) self-captured images, and (4) modified Hong Kong University Building Image Dataset (HBD) image set.

First, this research enhances the FireNet image set by covering additional firefighting assets and relabeling the images accordingly. The additional asset classes include fire door sign, fire hose reel, piping system, sprinkler, hidden fire equipment, emergency door release, emergency light, and firefighting lift switch. As a result, the total number of annotated instances increases from 2,154 to 2,878.

Another source of the Fire-ART dataset is the publicly accessible images. These images are systematically retrieved from the Google search engine (2025) using a custom-developed search API. There are two key criteria in the search process to ensure the relevance and usability of the result. First, the search employs target keywords related to fire equipment in building-specific contexts. For example, location-aware terms such as "emergency light in building," "fire pipes ceiling," and "fire door" are used instead of ambiguous and isolated queries like "fire door sign," to ensure that the target instances are captured in indoor scenes in a typical manner. Second, only images licensed under Creative Commons Zero (CC0) (Creative Commons, 2009) are selected to guarantee they are freely available for unrestricted use.

By employing the two search criteria, 3,043 images are downloaded from the Internet. However, the initial results included duplicated, low-resolution, and irrelevant images, which requires a rigorous filtering process to retain the quality of the dataset. For duplicated images, this research leverages Duplicate Cleaner (DigitalVolcano Software, 2022) to remove them automatically. Next, low-resolution and irrelevant images are manually screened and excluded. This process remains 468 high-quality, relevant images and 1,700 instances covering all the required classes.

The third source of the dataset consists of self-captured images collected by smartphones across different locations and settings to enhance the generalizability of the dataset. These images were photographed in various built environments, including shopping malls, airports, hotels, gym rooms, parking areas, restaurants, and office buildings, spanning multiple countries and regions, i.e., the United Kingdom, Mainland China, Macau SAR, France, the United States, and Greece. In this way, firefighting assets with region-specific design variations are incorporated in the Fire-ART dataset to ensure it has certain geographical diversity. Additionally, to address class imbalance, this self-capturing process targets to include firefighting assets that have relatively lower instance numbers in the above two sources, including sprinklers, emergency lights, and fire door signs. Consequently, this subset contributes 461 images and 1,142 instances to the Fire-ART dataset.

Last, this research also incorporates images from other built environment datasets for further expansion. Specifically, the Hong Kong University Building Image Dataset (HBD) is utilized (Wong et al., 2024) since it covers a variety of interior scenes (e.g., office, classroom, lab, study area, canteen, and pantry) from both historical and modern Eastern and Western-style buildings. A subset of 245 images, which focus on firefighting assets that are limited in the above three sources, is selected and labeled manually, resulting in 907 instances to constitute the Fire-ART dataset.

3.1.2   Firefighting asset labeling

In this research, images are labeled using the X-AnyLabeling tool (Wang, 2023) to delineate object boundaries with the assistance of Segment Anything model (SAM) (Kirillov et al., 2023). While SAM facilitates automated segmentation, it sometimes generates fragmented boundaries and includes extra irrelevant items. Therefore, manual checking and adjustment are required to ensure the firefighting asset instances are accurately annotated in the dataset.

Given that built environments are often complex and multiple assets may co-occur and visually overlap with each other, challenges exist and arise confusing labeling issues across different annotators. To make sure of the consistency of the dataset, this research sets several standard rules to guide the annotation process:

 (1) For hose reel cabinets, if multiple firefighting assets, such as extinguishers or standpipe

outlets, are visibly integrated within the cabinet, each asset is labeled individually (see (a) in Fig. 2). Otherwise, the cabinet is labeled as a single fire hose reel instance, as shown in (b) Fig. 2.

(2) Hose reels and hose pin racks are labeled as fire hose reel class due to their similar functions in fire suppression.

(3) To avoid duplication, fire exit signs with integrated emergency lighting functions are labeled as fire exit sign class, while the emergency light class is used for standalone emergency light fixtures without directional signage.

(4) For simplification, the components of piping systems, including pipes, outlets, and valves, are annotated under a single piping system class.

(5) In situations where firefighting assets are embedded into walls with no clear distinguishing features or visible cues to accurately determine their types, a separate hidden fire equipment class is introduced. This class includes items like hidden hose reels and extinguishers that are concealed behind panels or flush-mounted into walls. This hidden pattern makes the appearance of instances significantly differ from those visible assets. Therefore, this research separates them as different classes to avoid poor generalization and reduce misclassification errors (see (c) in Fig. 2).

(6) All firefighting assets are annotated at the instance level, except for the piping system class. For pipes, as their forms are various, sometimes intertwined, and easily affected by obstacles and joints, they are annotated semantically to simplify the labeling process and reduce ambiguity (see (d) in Fig. 2).

By following the above rules, three of the authors take part in the labeling of the images, followed by two rounds of cross-checking and refinement to ensure the quality of the dataset. Consequently, the Fire-ART dataset includes 15 firefighting assets, as shown in Fig. 3.

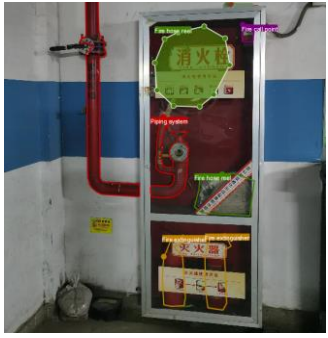

(a) Individual annotations for cabinets with multiple assets

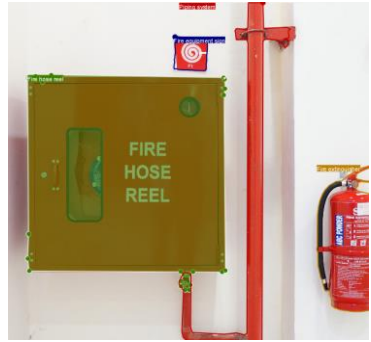

(b) Single annotation for cabinets with invisible assets

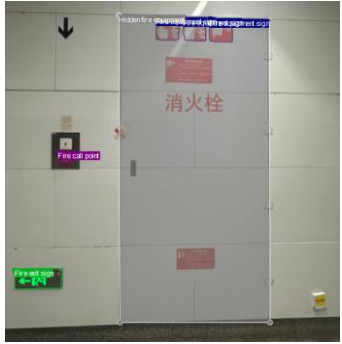

(c) Hidden fire equipment

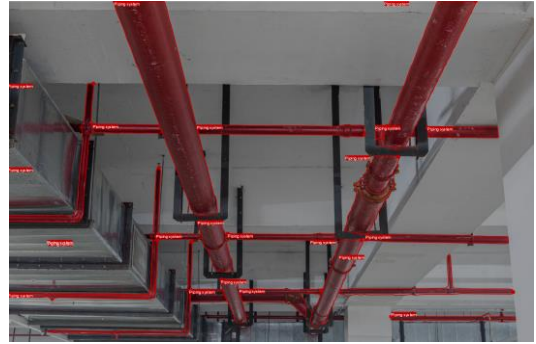

(d) Semantic annotations of piping systems

Fig. 2. Illustrations of annotation rules

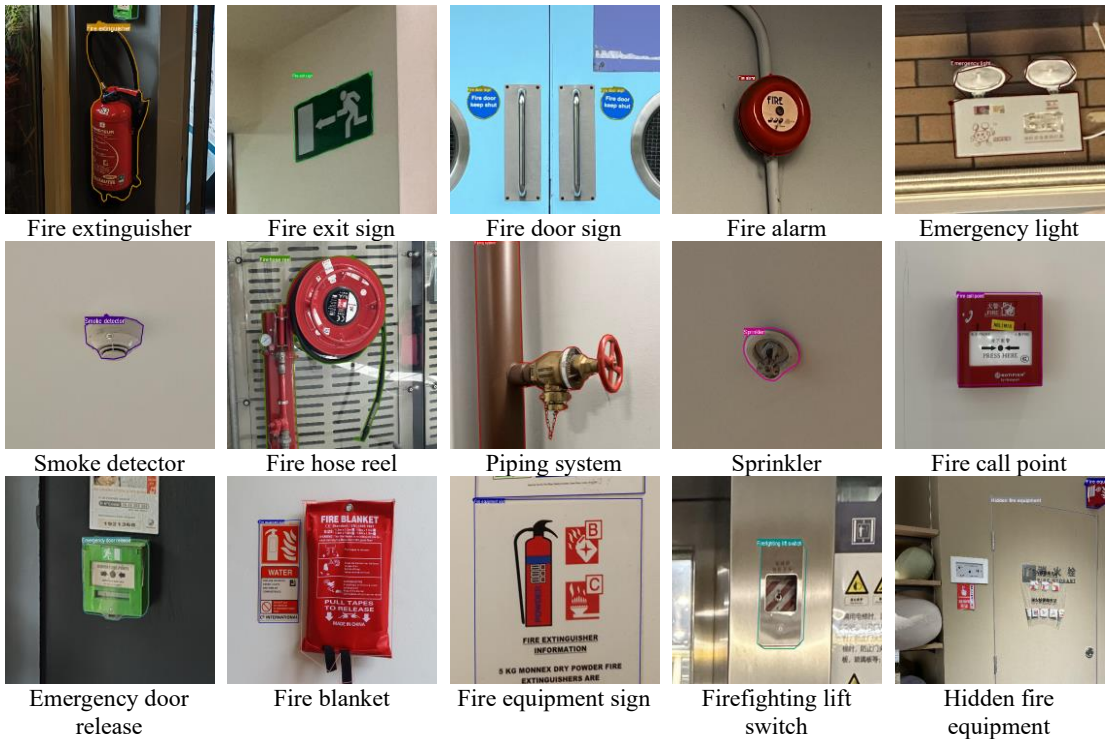

Fig. 3. Classes in the Fire-ART dataset

### 3.1.3 Data statistics

This section summarizes the statistics of the Fire-ART dataset. In total, the dataset comprises 2,626 images and 6,627 annotated firefighting asset instances. The number of images of each source is shown in Fig. 4. Moreover, Fig. 5 presents the instance numbers per class in the Fire-ART and the original FireNet datasets for direct comparison. The statistical results show that the dataset has an average of 441.8 instances per class and an average of 2.52 instances per image. Among all classes, the piping system has the highest number of instances (1,146), whereas the firefighting lift switch has the fewest instances (58). Approximately half of the classes contain over 400 instances. Despite targeted efforts to address class imbalance, it remains a challenge, particularly for emergency light and firefighting lift switch classes, due to their relatively rare occurrence in real-world environments.

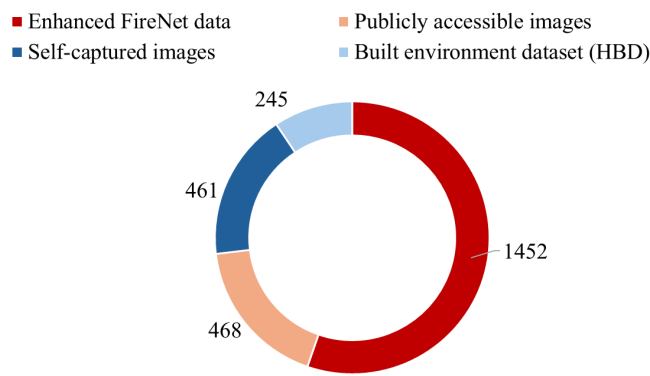

Fig. 4. Numbers of images from different sources

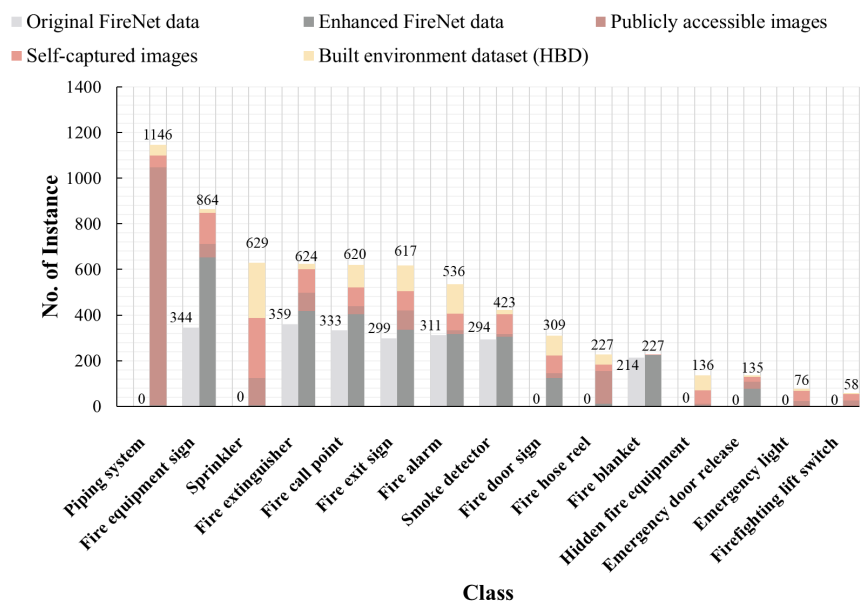

Fig. 5. Instance distribution per class between the Fire-ART and the original FireNet dataset

## 3.2 Image-based segmentation training

On the basis of the Fire-ART dataset, this research further leverages supervised learning techniques for firefighting assets recognition within images. Specifically, the off-the-shelf instance segmentation model, i.e., YOLOv8 (Jocher et al., 2023), is selected since it enables rapid inference while maintaining good segmentation accuracy. To conduct training, the Fire-ART dataset is split into a train, validation, and test subsets in line with a ratio of 0.70:0.15:0.15. As a result, the train, validation, and test subsets have 1,838, 393, and 395 images respectively.

The training is performed using a desktop computer, with AMD Ryzen 5 5600X CPU, NVIDIA GeForce RTX 4070 Ti GPU and 12GB Video RAM/GPU memory (VRAM). Consequently, the best-performing model with the least validation loss is selected for real-world deployment to predict instances of firefighting assets within images. Fig. 6 shows the predictions on the test subset images. The results on the test subset are summarized in Table 3.

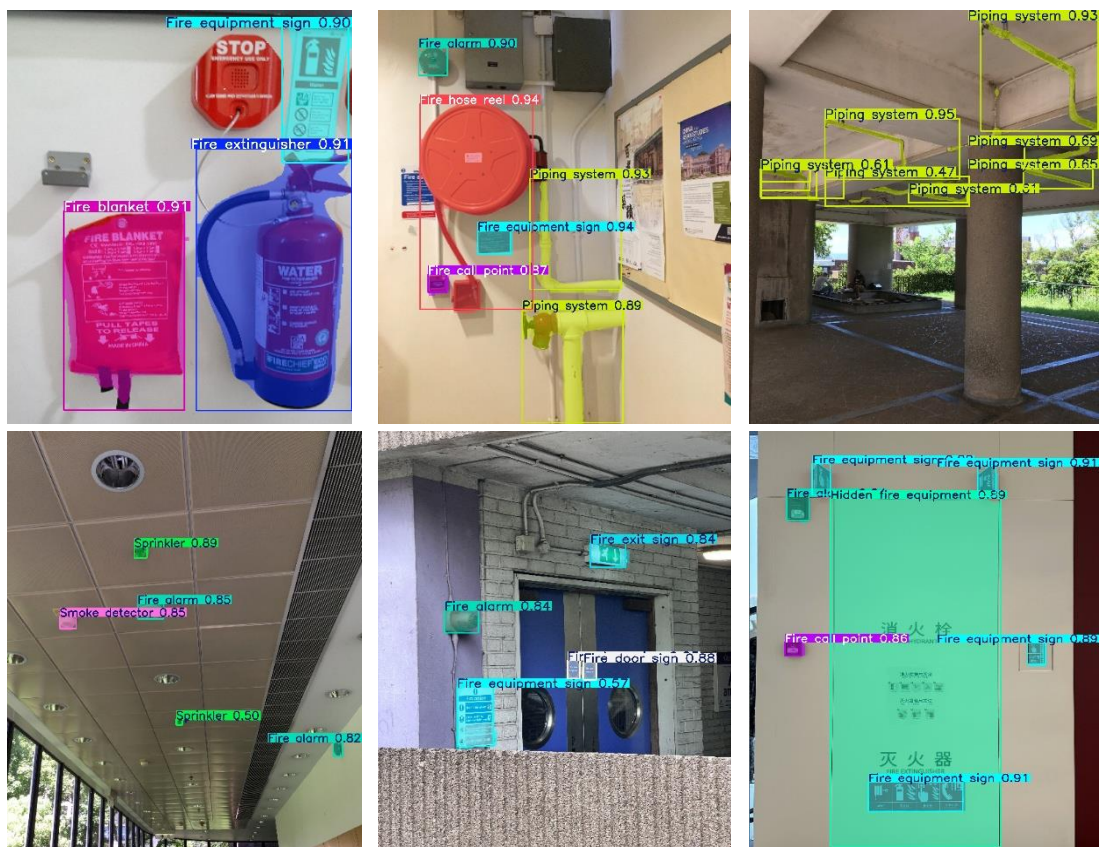

Fig. 6. Prediction results of the YOLOv8 model trained with the Fire-ART dataset

Table 3. Recognition accuracy results on the test set

| Class | Instances | Precision | Recall | mAP$^{50}$ | mAP$^{50-95}$ |
|---|---|---|---|---|---|
| Fire extinguisher | 82 | 0.891 | 0.841 | 0.888 | 0.576 |
| Fire exit sign | 102 | 0.863 | 0.784 | 0.848 | 0.517 |
| Fire door sign | 47 | 0.769 | 0.638 | 0.713 | 0.372 |
| Fire alarm | 74 | 0.888 | 0.859 | 0.92 | 0.59 |
| Emergency light | 14 | 0.647 | 0.571 | 0.617 | 0.498 |
| Smoke detector | 67 | 0.863 | 0.752 | 0.808 | 0.51 |
| Fire hose reel | 42 | 0.7 | 0.667 | 0.716 | 0.552 |
| Piping system | 161 | 0.695 | 0.325 | 0.404 | 0.221 |
| Sprinkler | 94 | 0.763 | 0.426 | 0.515 | 0.224 |
| Fire call point | 87 | 0.94 | 0.897 | 0.943 | 0.616 |
| Emergency door release | 33 | 0.894 | 0.455 | 0.585 | 0.383 |
| Fire blanket | 28 | 0.977 | 0.964 | 0.97 | 0.791 |
| Fire equipment sign | 113 | 0.864 | 0.876 | 0.893 | 0.672 |
| Firefighting lift switch | 10 | 0.873 | 0.4 | 0.474 | 0.419 |
| Hidden fire equipment | 20 | 0.881 | 0.743 | 0.851 | 0.766 |
| All | 974 | 0.834 | 0.68 | 0.743 | 0.514 |

The results indicate that the segmentation accuracy is moderately high, with a mAP$^{50}$ of 0.743 and a mAP$^{50-95}$ of 0.514. Specifically, the fire blanket (mAP$^{50}$ of 0.97), fire call point (0.943), fire alarm (0.92), fire equipment sign (0.893), and fire extinguisher (0.888) classes obtain relatively high accuracy, probably due to their consistent visual appearance and the availability of sufficient training data. By contrast, the piping system (0.404) class exhibits the lowest accuracy, which may be attributed to its complex and intertwined nature as well as the semantic-level annotation manner. In addition, the sprinkler (0.515), firefighting lift switch (0.474), and emergency door release (0.585) classes perform below average, which may be explained by the limited amount of training data and their relatively small size within the images.

**3.3 Reality capture and 3D semantic reconstruction**

3.3.1  Point cloud generation

Next, this research employs a 360-degree panoramic camera-based approach to conduct reality capture by collecting video data of interior scenes. The 360-degree camera is selected as it efficiently captures full coverage of a scene from all angles, containing abundant visual features in each shot. Compared to conventional cameras with limited fields of view, the 360-degree camera is advantageous in providing overlapping visual patterns for scene reconstruction and accelerating the capturing process. To perform reality capture using 360 cameras, several empirical principles are identified through pilot tests. These include maintaining a consistent walking speed, capturing at the highest resolution, stabilizing the camera stick vertically, avoiding movable objects during capture, and opening doors beforehand to maintain continuity of visual features when transitioning between multiple spaces. These principles, while simple, are the keys to ensure the success of point cloud reconstruction.

After reality capture, the videos are converted into image frames using an open-source media processing tool named FFmpeg (FFmpeg Developers, 2016). These image frames serve as the input for photogrammetric reconstruction. Next, an off-the-shelf software, Metashape (Agisoft, 2019), is utilized to perform structure-from-motion and multi-view stereo processes. In these processes, visual features are extracted and matched across overlapping image frames to generate a 3D point cloud structure and to estimate the camera pose trajectory. Fig. 7 shows the results of the point cloud reconstruction of an interior corridor scene.

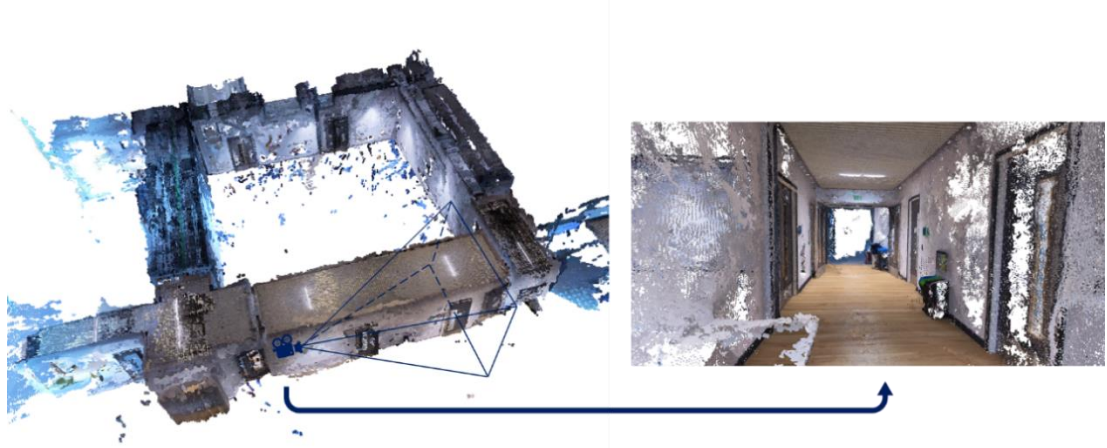

Fig. 7. Point cloud reconstruction result of an interior corridor

3.3.2　Cube-map conversion for image segmentation

The reconstructed point cloud consists of numerous points representing various interior elements, including building components, firefighting assets, and other irrelevant clutter. Hence, it is important to distinguish the interested point clusters from the others in order to correctly identify and localize the firefighting assets within the 3D scenes. To achieve this, the research adopts an indirect segmentation approach that first delineates firefighting assets as different classes in individual image frames, and then transfers these class labels to the 3D point cloud through camera projection. Compared to direct point cloud segmentation, this indirect segmentation approach is particularly effective for photogrammetry-based point clouds, as it leverages the rich visual information available in images as well as the abundance of image segmentation datasets. In addition, the inherent relationships between image planes, camera poses, and point cloud structures are utilized for camera projection to ensure the accuracy performance of the overall segmentation.

To conduct image segmentation of firefighting assets, this research employs the YOLOv8 model trained on the Fire-ART dataset. However, challenges arise due to the differences between normal images and 360-degree images. Specifically, images captured by 360-degree cameras are typically represented in an equirectangular form, which distorts the appearances of objects, making them unsuitable for direct input into the trained model. To address this issue, this research proposes a modified cube-map conversion to transform equirectangular images into rectilinear images. Conventionally, the cube-map conversion outputs six images representing the top, bottom, left, right, front, and back faces. Although this conventional conversion is capable of covering every aspect seamlessly, two primary issues exist and hinder the subsequent image segmentation process. First, firefighting asset instances may become fragmented across two separate face images, making them incomplete and hard to recognize. Second, the top and bottom face images appear unnatural since humans rarely capture images from these angles. Nevertheless, these images contain critical firefighting assets, such as sprinklers and smoke detectors mounted on ceilings. Therefore, it is important to retain the top and bottom face images while presenting them in a more natural manner. Fig. 8 (a) illustrates the two primary issues.

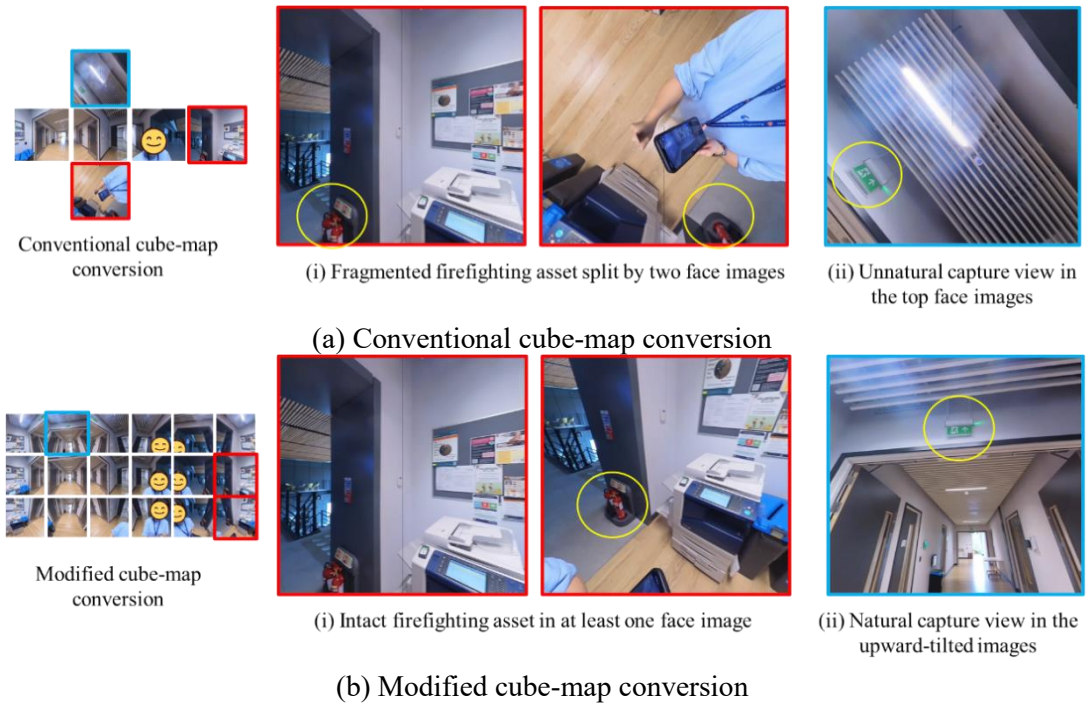

(b) Modified cube-map conversion
Fig. 8. Comparison of the conventional and the modified cube-map conversion approaches

To effectively tackle these two issues, this research modifies the cube-map conversion to produce a set of 18 rectilinear face images from each equirectangular image. First, six horizontal face images are generated, in which every two adjacent images would have a 30-degree overlap. This ensures that firefighting asset instances appear intact in at least one image (Fig. 8 (b)). Next, two additional sets of images are generated by tilting the camera angle upwards and downwards by 30 degrees from the horizontal orientation, each set containing six images. These tilted images extend the coverage to better capture ceiling-mounted or floor-level elements while preserving natural viewing angles. Fig. 8 displays a comparison of the tilted-up images from the modified conversion and the top face images from the conventional approach. The pseudocode script of the modified cube-map conversion is presented in Fig. 9.

**Modified cube-map conversion algorithm**

Input: *image_folder*, *output_folder*, *nb_splits*, *split_resolution*
Output: *split_images*

# Step 1: Define function to compute the rotation matrix for a given face
Define function `get_rotation_matrix`(*face_type*, *index*, *nb_splits*):
- Compute *yaw* angle, $yaw = (index / nb\_splits) * 2\pi - \pi$
- Determine *pitch* based on *face_type*:
  - If *face_type* is "top", set $pitch = \pi/6$    // 30° upward
  - Else if *face_type* is "bottom", set $pitch = -\pi/6$    // 30° downward
  - Else, set $pitch = 0$    // Horizontal view (no pitch)
- Compute rotation matrices:
  - *R_pitch* = rotation matrix from rotation vector $[pitch, 0, 0]$
  - *R_yaw* = rotation matrix from rotation vector $[0, yaw, 0]$
- Return the combined rotation matrix $R\_yaw * R\_pitch$

# Step 2: Generate face configuration list for cube map conversion
Initialize an empty list *faces_info*
For each index *i* from 0 to *nb_splits* - 1:
- Append ("horizontal", *i*, *nb_splits*) to *faces_info*    // Horizontal views
- Append ("top", *i*, *nb_splits*) to *faces_info*    // 30° upward views
- Append ("bottom", *i*, *nb_splits*) to *faces_info*    // 30° downward views

# Step 3: Load equirectangular image and define mapping parameters
List all image files in *image_folder* with supported extensions (.png, .jpg, .jpeg)
For each image file:
- Read the equirectangular image from *image_path*
- Extract image dimensions: *H* (height) and *W* (width)
- Set focal length factor $f = split\_resolution / 2$

# Step 4: For each face configuration, compute mapping arrays and generate rectilinear image
- For each (*face_type*, *split_index*, *nb_splits*) in *faces_info*:
  - Compute rotation matrix $R$ = `get_rotation_matrix`(*face_type*, *split_index*, *nb_splits*)
  - Initialize 2D arrays *map_x* and *map_y* of size (*split_resolution* * *split_resolution*)
  - For each pixel row *y* from 0 to *split_resolution* - 1:
    - For each pixel column *x* from 0 to *split_resolution* - 1:
      - Compute normalized coordinates:
        - $x\_norm = (x - (split\_resolution / 2)) / f$
        - $y\_norm = (y - (split\_resolution / 2)) / f$
        - $z\_norm = 1$
      - Form vector $V = [x\_norm, y\_norm, z\_norm]$
      - Rotate V: $V\_rotated = R * V$
      - Normalize *V_rotated* by dividing it by its magnitude
      - Convert *V_rotated* to spherical coordinates:
        - $lon = arctan2(V\_rotated[0], V\_rotated[2])$
        - $lat = arcsin(V\_rotated[1])$
      - Map spherical coordinates to equirectangular image coordinates:
        - $x\_src = (lon + \pi) / (2\pi) * W$
        - $y\_src = (lat + (\pi / 2)) / \pi * H$
      - Set $map\_x[y, x] = x\_src$ and $map\_y[y, x] = y\_src$
  - Remap the equirectangular image using *map_x*, *map_y* to create the rectilinear image
  - Save rectilinear image to *output_folder*
Return completion

Fig. 9. Pseudocode of the modified cube-map conversion algorithm

After conducting the cube-map conversion, the rectilinear face images are then performed predictions using the trained YOLOv8 model. As a result, the prediction masks are generated to delineate the boundaries of firefighting asset instances. During the process, a false-positive recognition issue is observed due to the presence of the surveyors in the 360-degree captures.

Specifically, the surveyors in the images are sometimes predicted as incorrect firefighting assets (e.g., fire hose reels). To address this, the research employs another off-the-shelf instance segmentation model pre-trained on the COCO dataset to rapidly detect and segment human regions. These human segmentation masks are then applied inversely to rectify incorrect predictions in the original images, thereby substituting the misclassification and enhancing the segmentation results, as shown in Fig. 10.

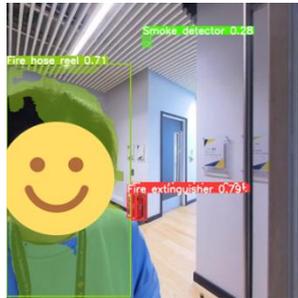
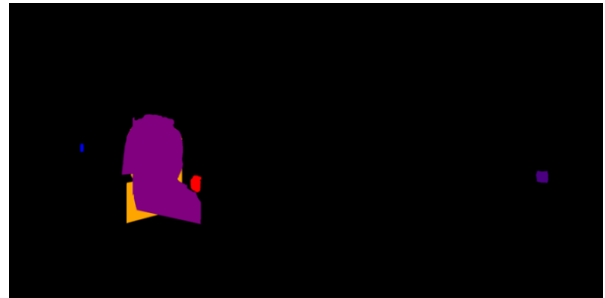

(a) Mis-categorization of human as fire hose reel  (b) Incorrect mask

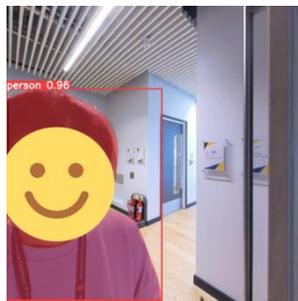
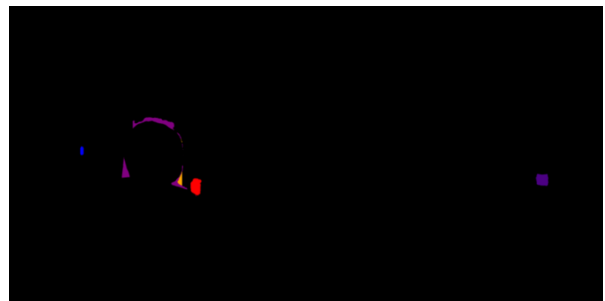

(c) Human segmentation  (d) Mask after inverse operation

Fig. 10. Inverse operation for human-related mis-categorization

Subsequently, these corrected rectilinear masks are further merged into a unified equirectangular mask that is used for spherical camera projection. Since each capture produces 18 prediction masks with partial overlaps, it is crucial to resolve the possible prediction conflicts across different face masks. Therefore, this research proposes a class priority voting (CPV) algorithm, specifically designed to handle two conflict situations: (1) a pixel region identified as a firefighting asset in certain images but classified as background in others, and (2) a pixel region identified as different firefighting asset classes across multiple images. For situation (1), the CPV algorithm prioritizes firefighting asset classes over the background class to enhance their likelihood of being recognized. For situation (2), the algorithm counts the occurrence of each interested class across the overlapping masks and further assigns the class with the highest frequency to the conflicting pixel regions. Fig. 11 presents the pseudocode of the CPV algorithm. The workflow of creating equirectangular mask is presented in Fig. 12.

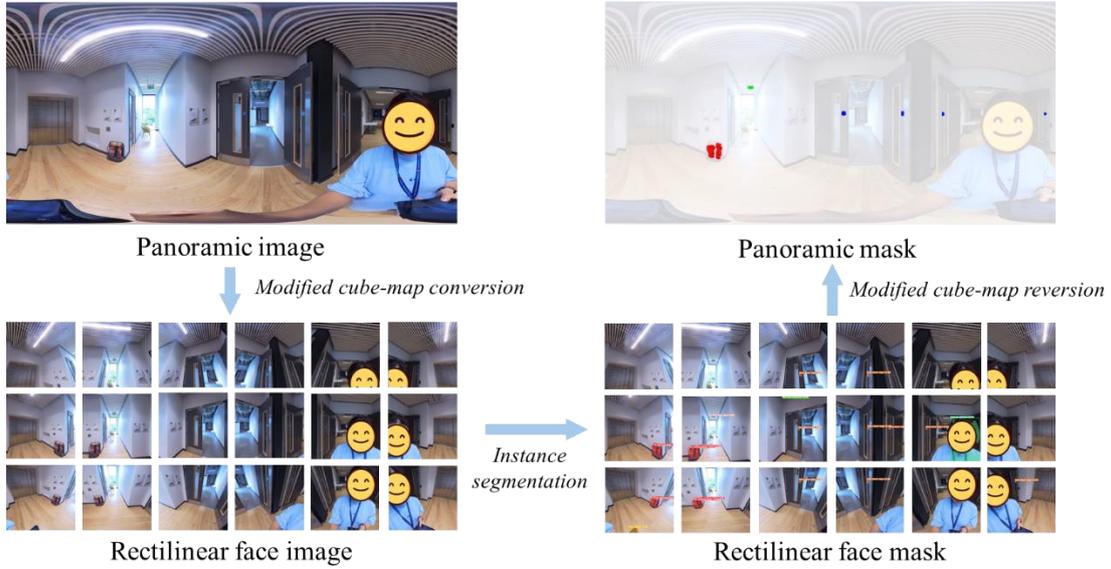

Fig. 11. Workflow of equirectangular mask creation

---

**Class priority voting algorithm**

Input: $masks\_list$, $nb\_splits$, $H$, $W$, $num\_classes$
Output: $equirectangular\_mask$

Create an $nb\_splits \times H \times W$ array storing the rectilinear masks
| $masks\_array = stack(masks\_list)$
Flatten the masks array
| $masks\_flat = reshape(masks\_array, (H \times W, nb\_splits))$
Create an array to store the final pixel classes
| $final\_labels$ = array of zeros with length $(H \times W)$
For each $index$ from 0 to $H \times W - 1$:
    Retrieve the class votes across all individual rectilinear masks
    | $votes = masks\_flat[index]$
    Initialize a counter for all classes
    | $counts$ = array of zeros with length $num\_classes$
    For each $c$ in $votes$:
    | $counts[c] += 1$
    Sort classes by descending vote count
    | $sorted\_ids = argsort(counts, descending = True)$
    Assign the most-voted class:
    | $final\_labels[index] = sorted\_ids[0]$
    If the assigned class is background, re-assign to the second-most-voted class:
    | $final\_labels[index] = sorted\_ids[1]$
Generate equirectangular mask $equirectangular\_mask = reshape(final\_labels, (H, W))$
Return $equirectangular\_mask$

Fig. 12. Pseudocode of the CPV algorithm

### 3.3.3 Camera projection

Based on the unified equirectangular masks, spherical camera projection is implemented to transfer the class information from the masks onto the 3D point cloud structure. The mechanism involves mapping the 3D point cloud points onto the image spherical surfaces, where each 3D point corresponds to a 2D pixel in the equirectangular image. By conducting the 3D-to-2D transformation, the classification labels of the pixel regions can then be assigned back to the 3D point cloud points accordingly.

Conventionally, the spherical camera model projects points onto a spherical surface by tracing rays from each point directly toward the camera center regardless of distance. However, in indoor scenarios, this projection method can lead to a penetration issue, where points behind walls or occlusions could be incorrectly classified. To mitigate the error, this research introduces a radius-based threshold to set the effective projection range. Specifically, only points $p_i$ within a predefined radius $r$ around the camera position $c_j$ are considered for projection, while points located beyond this radius are excluded. In this way, the projection would prioritize nearby point cloud objects, which are generally presented clearer and have better segmentation accuracy. Equations (1) – (3) illustrate the radius-based spherical projection mechanism. Fig. 13 shows the result of spherical camera projection of an equirectangular mask with radius thresholding.

$$\boldsymbol{P}_r = \left\{\boldsymbol{p}_i \middle| \|\boldsymbol{p}_i - \boldsymbol{c}_j\|_2 \leq r\right\} \quad (1)$$

$$\begin{cases} u_i = 0.5w + f \times \tan^{-1}(x_i/z_i) \\ v_i = 0.5h + f \times \tan^{-1}\left(y_i/\sqrt{x_i^2 + z_i^2}\right) \end{cases}, \quad \text{where } \boldsymbol{p}_i = (x_i, y_i, z_i) \in \boldsymbol{P}_r \quad (2)$$

$$f = w/(2\pi) \quad (3)$$

where $\boldsymbol{P}_r$ represents the set of points projected from the camera location $c_j$ within the radius range $r$, $(x_i, y_i, z_i)$ denotes the 3D coordinates of the cloud point $\boldsymbol{p}_i$, $h$ and $w$ is the height and width of the panoramic image, $f$ is the focal length in pixels, and $(u_i, v_i)$ corresponds to the pixel coordinate of the cloud point $\boldsymbol{p}_i$.

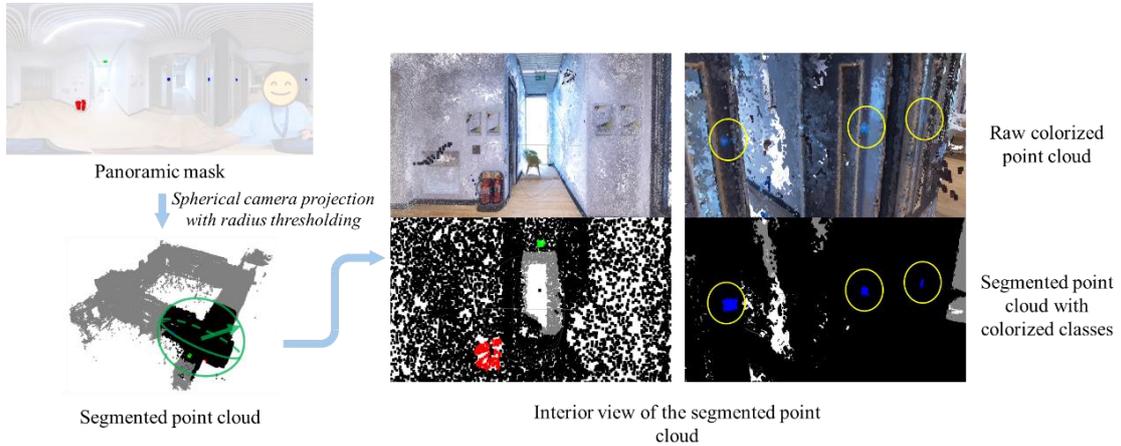

Fig. 13. Result of spherical camera projection with radius thresholding

Given that the entire indoor scenes are captured through a series of frame images, with each image covering only a partial area, the spherical projection is performed iteratively across all frames and accumulates complete segmentation results. Among the adjacent frame images, their coverage would have certain overlaps, in which inconsistent segmentation may occasionally occur due to the incorrect image prediction. To handle this inconsistency, this research employs a weighted majority voting approach adapted from (Wong et al., 2025). The weighted majority voting first counts the occurrence frequency of each semantic class across different projected images, and then multiplies these frequencies by predefined class-specific weights. The class with the highest weighted score is ultimately assigned as the final classification result. Specifically, the class-specific weights are introduced to enhance the recognition likelihood of firefighting assets over the background classes, where the weight values are determined through trial and error to retain the adaptability to different indoor scenarios. Consequently, the final classification of the 3D point cloud is collectively determined by considering all the projection results from different capture locations, resulting in a more robust and accurate segmentation performance in comparison to individual image predictions.

**3.4 BIM enrichment**

To effectively enrich the BIM model with detailed firefighting asset information, accurate alignment between the photogrammetric point cloud and the BIM model is critical. Challenges exist in the alignment process as the photogrammetric point cloud and the BIM model have two distinct coordinate systems. The traditional point cloud alignment approaches such as iterative closest points and normal distributions transform are not directly suitable given that the photogrammetric point cloud and the BIM model are from heterogeneous sources. In addition, since the photogrammetric point cloud may only capture portions of indoor scenes, the alignment requires partial-to-whole registration and thus increases the complexity, which remains an open research area deserving comprehensive investigation.

Correspondingly, this research proposes a semi-automated alignment approach. First, the fundamental BIM components, including floors, ceilings, walls, doors, and windows, are extracted to form polygonal surface geometries based on the IFC 4.3 data schema (2024). Then, the IFC model is converted into a synthetic point cloud. Random point sampling is then conducted within these polygon boundaries. The sampling process is iteratively performed among surfaces of different components, where the sampling point numbers are set in line with the surface areas to ensure the uniformity of cloud points. As a result, the synthetic IFC-BIM point cloud is yielded and ready for registration.

To align the synthetic with the photogrammetric point clouds, the research utilizes CloudCompare (2024) to perform pairwise point selection. This process involves manually identifying matching point pairs between the two point clouds to compute a transformation matrix for registering the photogrammetric point cloud into the BIM coordinate system. Following the alignment, this research further conducts clustering to separate individual instances of firefighting assets within the segmented point cloud. Specifically, the density-based spatial clustering of applications with noise (DBSCAN) algorithm (Ester et al., 1996) is employed to differentiate clusters based on their spatial distances and densities. For each cluster, the central position coordinate and the asset type are obtained to represent the associated instance. Based on the extracted data, this research enriches the BIM model to include all the identified firefighting assets. A Revit family library called Glodon digital component dock (2024) is utilized to enable rapid modeling using predefined firefighting asset families. Eventually, the BIM model is augmented with critical firefighting asset information, enhancing its capabilities to support different applications such as facility management and emergency planning.

## 4. Case study and results

### 4.1 Testbed background

In this research, two experiments are conducted using real-world corridor spaces of two institutional buildings at the University of Cambridge (case study one) and University of Macau (case study two) respectively. The corridor spaces are selected since they are typical indoor environments that is often mounted and equipped with various firefighting assets. The floor plans of two case studies are provided in Fig. 14. The testing area in case study one is 136.76m$^2$ and 394.06 m$^2$ for case study two. Representative firefighting assets from case study one and case study two are illustrated in Fig. 15 to highlight their specific features and visual characteristics.

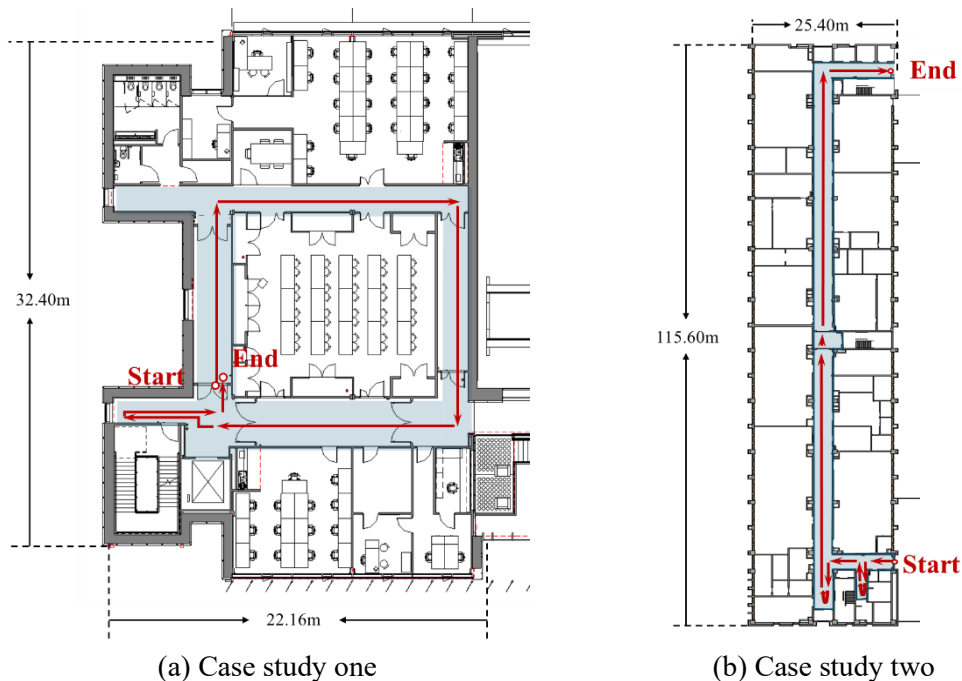

(a) Case study one            (b) Case study two
Fig. 14. Layouts and capture paths of the two case studies

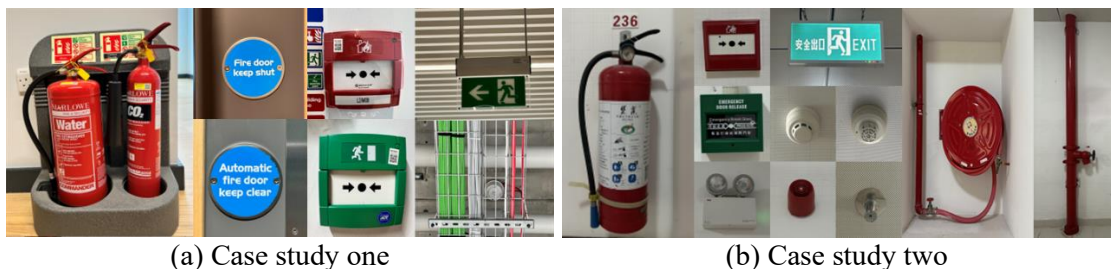

(a) Case study one            (b) Case study two
Fig. 15. Characteristics of firefighting assets in the two case studies

The testbed of case study one is a close-loop corridor on the first floor of the Civil Engineering Building at the University of Cambridge. The corridor is featured a standardized indoor configuration of 38 pieces of fire safety equipment, comprising six types of firefighting assets:

fire extinguisher, fire exit sign, fire door sign, smoke detector, fire call point, and emergency door release. In current practice, the as-is BIM model of this testbed integrated architectural, structural, and MEP systems with a level of development (LOD) of 300, while firefighting equipment was not included in the model.

To validate the robustness of the proposed approach under differing conditions, the second case study was conducted in a corridor on the third floor of E11 building at the University of Macau. This site presented a different configuration compared to case study one, particularly in the appearance, type variety, and density of installed firefighting equipment (Fig. 15). The corridor contained ten types of firefighting assets: fire extinguisher, fire exit sign, fire alarm, emergency light, smoke detector, fire hose reel, piping system, sprinkler, fire call point, and emergency door release (124 pieces in total). The as-is BIM model included basic architectural elements with a LOD of 200.

**4.2 Implementation**

The two case studies were conducted following a similar data acquisition and processing workflow. A detailed comparison of their configurations is provided in Table 4. The videos for case study one and case study two were recorded using Insta360 X3 and Insta360 X4 cameras, respectively, with durations of 4min 16s and 4min 36s. In both cases, the camera was handheld by the two of the authors, positioned at approximate heights of 1.6 m and 1.75 m.

The capturing and reconstruction workflows followed the proposed methodology. First, the videos were processed to extract 770 and 830 panoramic images for case study one and case study two, respectively, at a sampling rate of three frames per second. These panoramic images were subsequently converted into 13,860 and 14,940 rectilinear face images for the two cases using the OpenCV library (Bradski, 2000), which were imported into Metashape for point cloud reconstruction. Second, the face images were predicted using the trained YOLOv8 model based on the Fire-ART dataset. The face images with the labels of firefighting assets were then converted into semantic masks. These masks were merged to conduct spherical camera projection to assign different firefighting asset classes to the cloud points. As a result, the segmented point cloud was acquired, in which the locations and numbers of various firefighting assets were calculated via point cloud clustering. Finally, these firefighting asset data were leveraged to enrich the BIM models by aligning the point cloud to the IFC data model as well as inserting firefighting asset objects in line with the calculation results.

Table 4. Configurations of two case studies

| **Implementation** | **Case study one** | **Case study two** |
| --- | --- | --- |
| Equipment | Insta360 X3 | Insta360 X4 |
| Camera height | 1.6 m | 1.75 m |
| Video resolution | 3840 × 1920 pixels | 5760 × 2880 pixels |
| Video length | 4min 16s | 4min 36s |
| No. of panoramic images | 770 | 830 |
| No. of rectilinear face images | 13,860 | 14,940 |

### 4.3 Results

For evaluation, the ground truth models of the two testbeds were created by manually measuring the locations of firefighting assets and inserting the correct asset types into the as-is architectural BIM models. Next, several recognition and localization metrics were utilized to quantify the results against the ground truth. Specifically, precision ($p$), recall ($r$), and F1-score ($F1$) were used to evaluate recognition accuracy, while the distance errors ($d$) between the ground truth and the true-positive assets were calculated to assess localization accuracy. The calculation of these metrics is illustrated in Equations (4) – (7).

$$p = TP/(TP + FP) \tag{4}$$
$$r = TP/(TP + FN) \tag{5}$$
$$F1 = (2 \times p \times r)/(p + r) \tag{6}$$
$$d = \sum_{i=1}^{|TP|} \|loc_i^{pred} - loc_i^{gt}\|_2 / |TP| \tag{7}$$

where $TP$, $FP$, and $FN$ are the numbers of true-positive, false-positive, and false-negative instances, $loc_i^{pred}$ and $loc_i^{gt}$ denote the predicted and ground truth 3D coordinates of the $i$-th true positive instance, and $\|\cdot\|_2$ represents the Euclidean norm.

4.3.1   Results of case study one

The spatial distribution of recognized instances and their comparison with ground truth instances on the BIM floor plans are shown in Fig. 16. Overall, the model achieved an average precision of 82%, an average recall of 67% and an average F1-score of 73%, indicating a moderately high level of recognition performance. In addition, the distance error between the ground-truth and the true-positive instances was 0.620 m (see Table 5). The results indicate that recognition accuracy varied significantly across different types of firefighting assets. For instance, the fire call point attained 100% in both precision and recall rates, demonstrating that all instances of this asset were accurately recognized. Emergency door releases, exit signs, and door signs exhibited similarly high recognition metrics, attributable to their high visibility.

Several critical challenges also emerged in case study one. First, the fire extinguisher asset yielded a recall of only 50%. It is because the extinguishers appeared in pairs in the real-world environment, while the point cloud clustering result failed to distinguish them individually due to the close spatial distance, resulting in a lower detection rate (see Fig. 17 (a)). Second, one exit sign was falsely detected as negative because of the dark environment and far distance (see Fig. 17 (b)). Third, smoke detectors were obscured by ceiling decoration bars or pipelines. As shown in Fig. 17 (c), both pipelines and smoke detectors were mounted along the central axis of the corridor. The data collection trajectory followed a similar horizontal path, which caused visual occlusion and led to the false negatives of the eight smoke detectors. In the test area of case study one, there were four motion sensors installed on the ceiling and possessed a similar appearance to smoke detectors. Therefore, these four sensors were falsely detected as smoke detectors, leading to a complete failure in accurately detecting any smoke detectors. This results in zero values for precision, recall, and F1-score metrics, with no data for localization accuracy computation. Another challenge arose from the photogrammetric reconstruction process. During reconstruction, errors may occur due to incorrect matching or triangulation uncertainties. These issues gradually compromised geometric accuracy, as evidenced by the deviations in the reconstructed positions of identical objects captured at the beginning and end of the route. Consequently, the accumulated reconstruction errors adversely impacted localization accuracy.

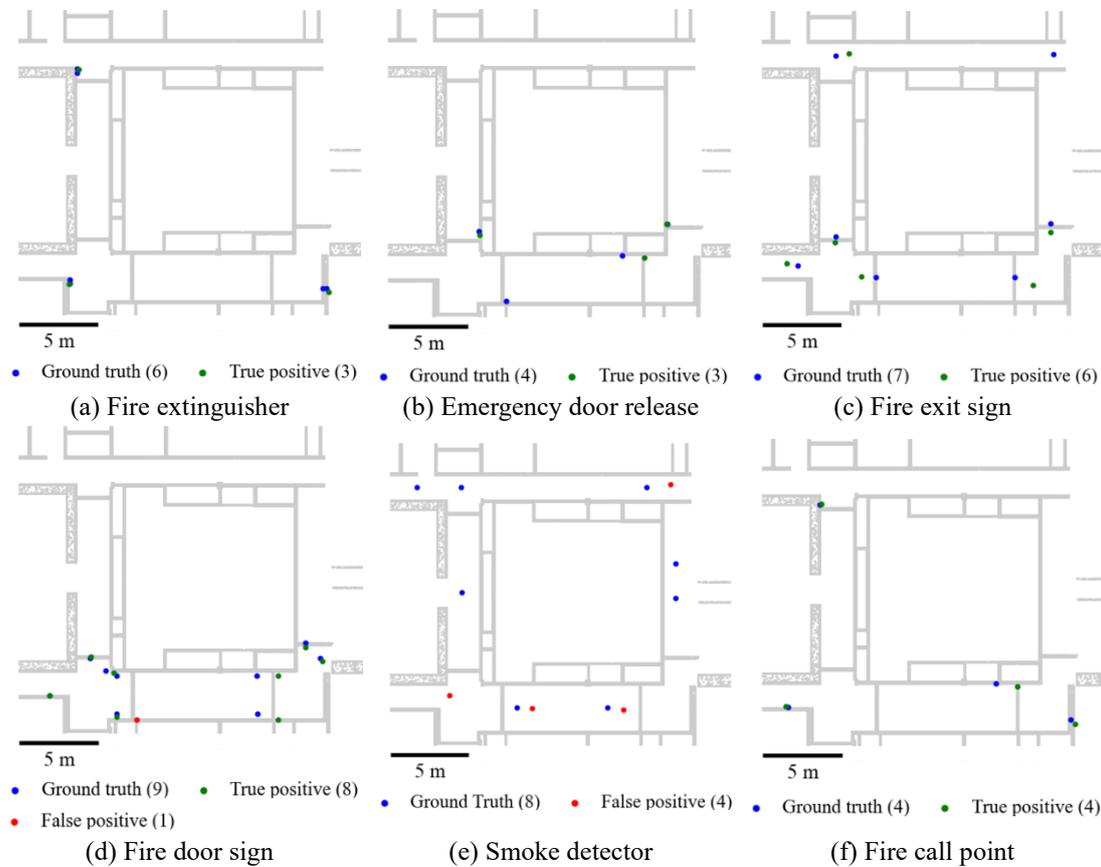

Fig. 16. Comparison between ground truth and results in case study one

Table 5. Results of case study one

| Firefighting asset | GT | TP | FP | FN | Precision | Recall | F1 | Distance error (m) |
|---|---|---|---|---|---|---|---|---|
| Fire extinguisher | 6 | 3 | 0 | 3 | 100% | 50% | 67% | 0.377 |
| Emergency door release | 4 | 3 | 0 | 1 | 100% | 75% | 86% | 0.620 |
| Exit Sign | 7 | 6 | 0 | 1 | 100% | 86% | 92% | 0.949 |
| Door sign | 9 | 8 | 1 | 1 | 90% | 90% | 90% | 0.525 |
| Smoke detector | 8 | 0 | 4 | 8 | 0% | 0% | 0% | - |
| Fire call point | 4 | 4 | 0 | 0 | 100% | 100% | 100% | 0.631 |
| Average | - | - | - | - | 82% | 67% | 73% | 0.620 |

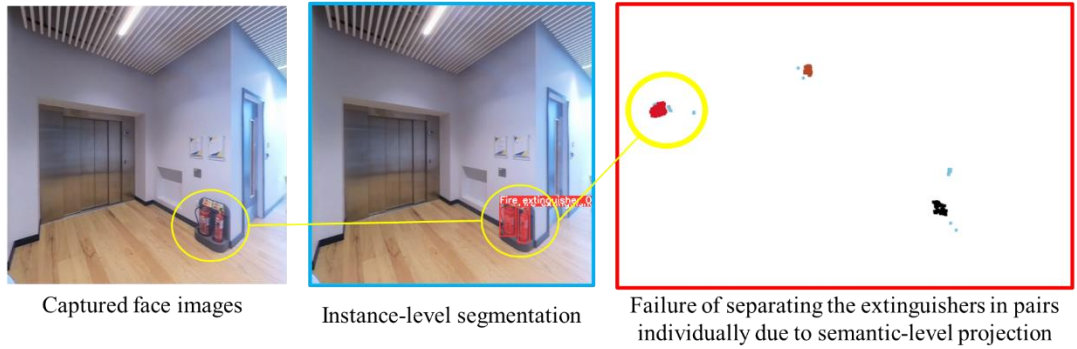

(a) False detection of fire extinguisher

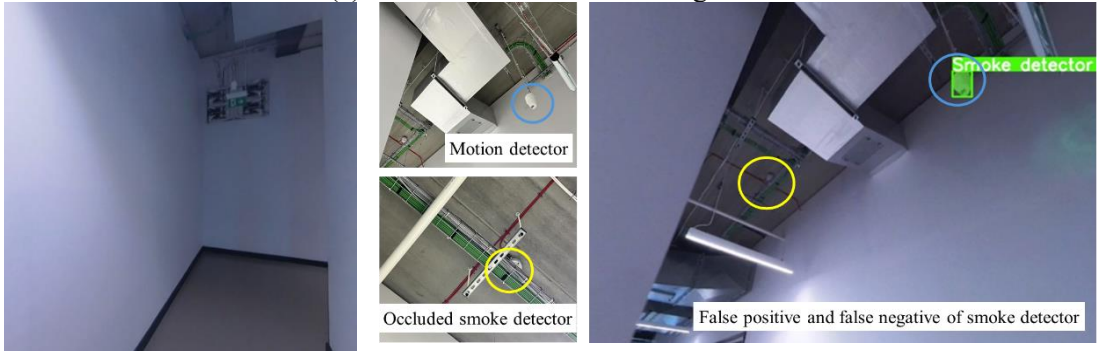

(b) False detection of fire sign in the dark environment

(c) False detection of smoke detector

Fig. 17. Examples of false detection in case study one

### 4.3.2 Results of case study two

The results of case study two are presented in Fig. 18 and Table 6, where the average precision, recall, and F1-score of the proposed approach are 94%, 84%, and 88%, respectively, indicating a robust recognition performance. In particular, several fire safety assets showed high recognition accuracy. Fire extinguishers, piping systems, and fire call points attained 100% precision and recall rates. Fire hose reels and fire alarms achieved 100% precision with recall rates exceeding 85%. Moreover, sprinklers reached 100% precision and 93% recall rates. It is noteworthy that the performance of recognizing these ceiling-mounted assets is improved considerably. Our prior research (Qiao et al., 2025), which only utilized horizontal rectilinear face images for asset recognition, demonstrated less satisfactory performance, with 100% precision and 56% recall. By contrast, the proposed approach modifies the cube-map conversion and incorporates those upward-tilted images, which improve the visibility and detection of the ceiling-mounted components.

Despite the promising results, the proposed approach also exhibited several limitations in detecting emergency door release, smoke detector, and occluded assets. Specifically, emergency door releases had the lowest recall rate (29%), with particularly poor recognition performance near corridor endpoints. This may be attributed to the location nature of emergency door releases in the scene. They were placed near doorways and thus were easily obscured by door leaves and walls during scene capture (Fig. 19 (a)). For smoke detectors, it achieved only 59% recall rate due to multiple factors. Further investigation revealed that the trained segmentation model could only recognize one type of smoke detector while missing the other type (Fig. 19

(b)), likely because the other type was underrepresented in the Fire-ART dataset. Moreover, as the smoke detectors were mounted on the ceiling near light fixtures, the lens flare effect occurred and adversely affected image capture and segmentation performance (Fig. 19 (b)). Lastly, the recognition of smoke detectors located at the end of the capture trajectory was reduced, largely due to a lower number of frames capturing sufficient visual features for identification. Additionally, occlusions also caused missed detections, including one fire exit sign obscured by a water dispenser (Fig. 19 (c)). Finally, the localization accuracy was slightly better than that of case study one, with an average distance error of 0.428 m. The error was mainly influenced by the uncertainties in the photogrammetric reconstruction and semi-automated point cloud-to-BIM alignment processes.

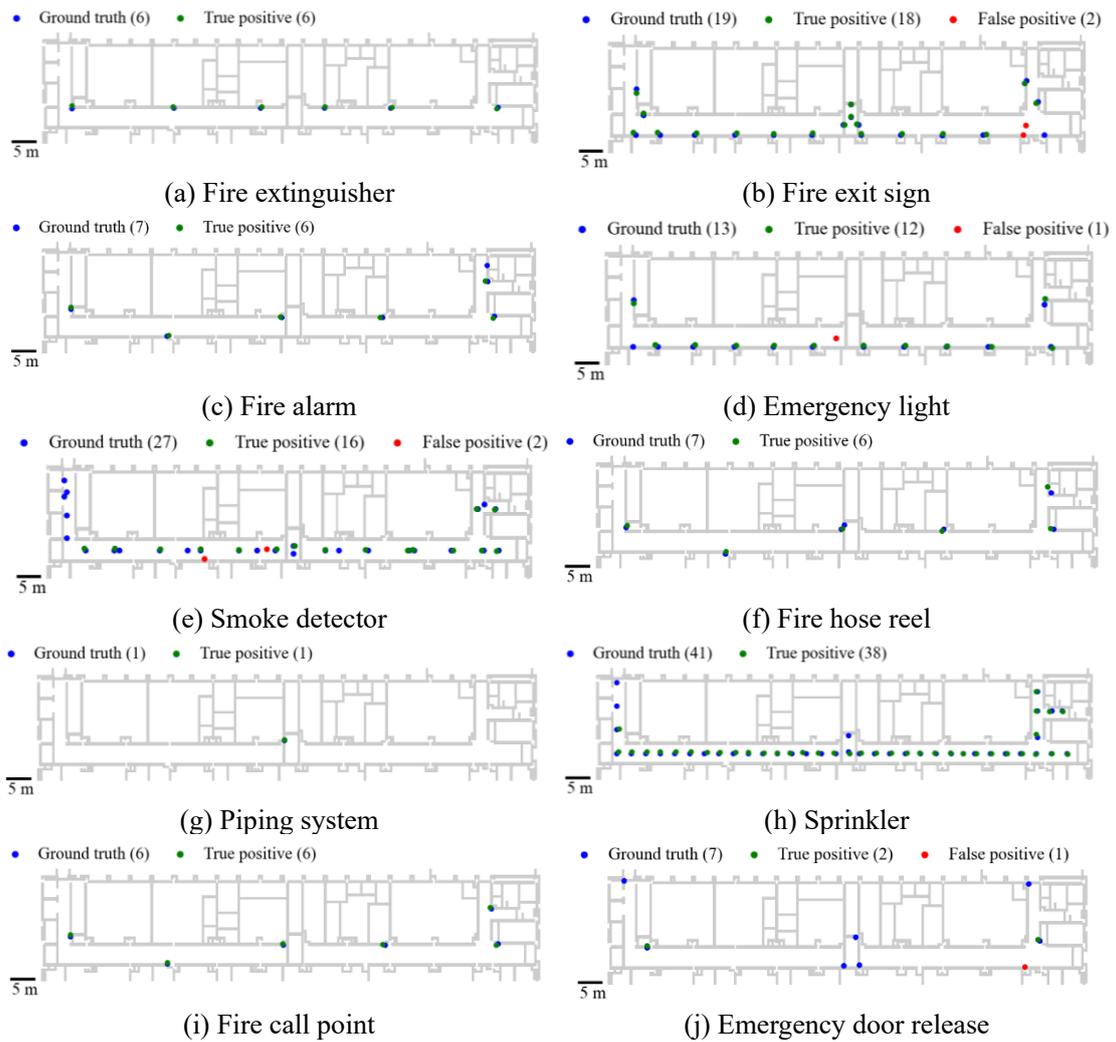

Fig. 18. Comparison between ground truth and results in case study two

Table 6. Results of case study two

| Firefighting asset | GT | TP | FP | FN | Precision | Recall | F1 | Distance error (m) |
|---|---|---|---|---|---|---|---|---|
| Fire extinguisher | 6 | 6 | 0 | 0 | 100% | 100% | 100% | 0.357 |
| Fire exit sign | 19 | 18 | 2 | 1 | 90% | 95% | 92% | 0.465 |
| Fire alarm | 7 | 6 | 0 | 1 | 100% | 86% | 92% | 0.419 |
| Emergency light | 13 | 12 | 1 | 1 | 92% | 92% | 92% | 0.500 |
| Smoke detector | 27 | 16 | 2 | 11 | 89% | 59% | 71% | 0.322 |
| Fire hose reel | 7 | 6 | 0 | 1 | 100% | 86% | 92% | 0.604 |
| Piping system | 1 | 1 | 0 | 0 | 100% | 100% | 100% | 0.360 |
| Sprinkler | 41 | 38 | 0 | 3 | 100% | 93% | 96% | 0.374 |
| Fire call point | 6 | 6 | 0 | 0 | 100% | 100% | 100% | 0.388 |
| Emergency door release | 7 | 2 | 1 | 5 | 67% | 29% | 40% | 0.423 |
| Average | - | - | - | - | 94% | 84% | 88% | 0.428 |

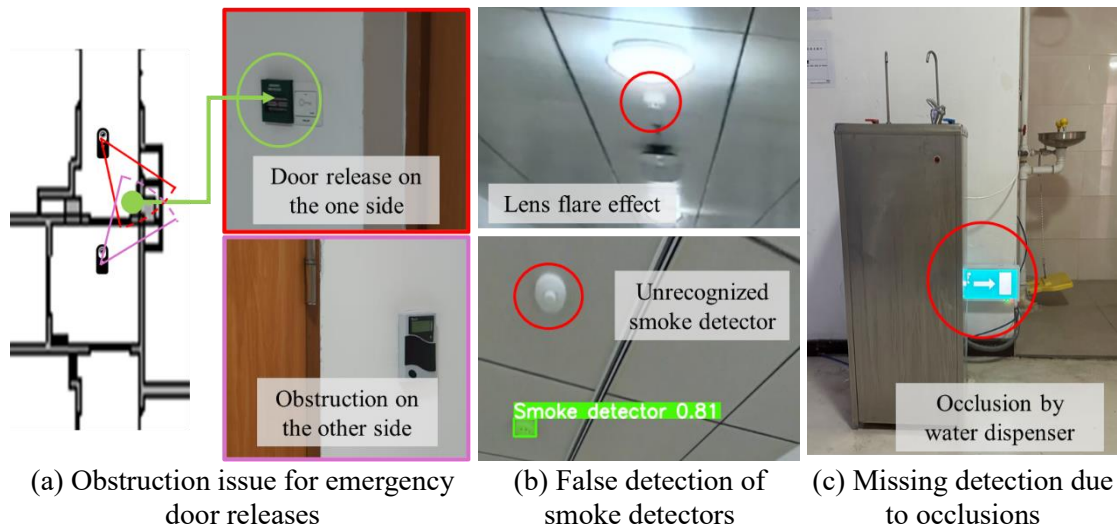

(a) Obstruction issue for emergency door releases    (b) False detection of smoke detectors    (c) Missing detection due to occlusions

Fig. 19. Examples of false detection in case study two

## 5. Discussion

Accurate and up-to-date information on indoor fire protection assets plays a significant role in both daily facility management and emergency response. However, existing literature lacks comprehensive investigations that present holistic approaches for semantic information modeling of firefighting assets. The key obstacles lie in the limited availability of firefighting asset datasets and the time- and resource-intensive nature of data collection and processing. To address these challenges, this research develops the Fire-ART dataset and proposes a cost-effective panoramic image-based 3D reconstruction approach to enrich firefighting asset information within BIM models. The implications of this research, challenges encountered in various stages, and the limitations and future work are elucidated as follows.

## 5.1 Characteristics of the Fire-ART dataset

This research proposes an extensive and publicly accessible dataset, Fire-ART, for firefighting asset recognition. The Fire-ART dataset not only expands the existing FireNet dataset by including extra asset classes, but also incorporates diverse resources through a systematic collection of publicly available images, self-captured photographs, and other built environment datasets. As a result, the Fire-ART dataset comprises 2,626 images, containing 6,627 instances of firefighting assets across 15 essential classes.

The proposed Fire-ART dataset has several critical features that distinguish it from existing efforts. First, it includes additional fundamental assets that are rarely available in prior datasets, such as sprinklers, piping systems, emergency door releases, and hidden fire equipment. Second, the dataset provides almost instance-level annotations (except for piping systems), which enables precise boundary delineation and instance counting compared to existing object detection or semantic segmentation datasets. Third, the dataset is curated to cover diverse indoor environments (e.g., shopping malls, offices, and airports) across various geographical regions, including the United Kingdom, Mainland China, Macau SAR, France, the United States, and Greece. In this way, different appearances and language characters used in firefighting assets across different countries or regions can be captured to enhance the generalizability of the dataset. Last, Fire-ART represents the most extensive publicly available dataset to date for supporting model training and benchmarking in firefighting asset recognition.

Compared to widely used FireNet dataset, the Fire-ART dataset exhibits an 80.9% increase in the number of images with over a twofold expansion in labeled asset instances. Additionally, the average instance number per image and per class in Fire-ART are 2.52 and 441.8, respectively, both of which are higher than the counterparts of FireNet (1.48 instances per image and 269.3 instances per class). These improvements indicate that Fire-ART contains richer content within individual images and more sufficient data for each asset type. As a result, it serves as a valuable resource for research and practical applications in advancing more automated and accurate recognition of firefighting assets.

## 5.2 Semantic reconstruction for BIM model enrichment

Apart from the Fire-ART dataset, this research introduces a novel panoramic image-based 3D reconstruction approach for semantic BIM enrichment of firefighting assets. Particularly, the proposed approach has three distinct benefits. First, it leverages photogrammetry-based reconstruction by making use of panoramic images captured via 360-degree cameras. Specifically, panoramic images offer details of visual features and empower full coverage of indoor environments. These advantages are instrumental for accurately reconstructing scenes to compute the relative locations of firefighting assets while also improving capture efficiency and spatial completeness.

Second, this research utilizes and further refines cube-map conversion to enhance recognition performance. The cube-map conversion is useful in transforming equirectangular images into rectilinear images to mitigate distortion issues. In conventional research, one panoramic image is divided into six rectilinear images representing the top, bottom, left, right, front, and back faces. However, such a conversion incurs unnatural viewpoints and fragmented assets across different face images. To overcome these limitations, a modified cube-map conversion is proposed to convert each equirectangular image into 18 face images with natural viewing angles and intact representations of asset instances. Subsequently, a reversion process is conducted to merge the predicted face masks into a unified equirectangular mask for semantic projection. In this way, the modified cube-map conversion offers two primary advantages: (1) it effectively resolves the representational discrepancies between panoramic imagery and conventional normal image datasets, thereby maximizing the utilization of existing recognition models, and (2) it establishes the mapping relationship between the equirectangular and the rectilinear images, which forms the basis for future development of comprehensive 360-degree image datasets.

Last, this research employs spherical camera projection to map between semantic information from image prediction and 3D point cloud structure. Particularly, radius thresholding is developed to address the penetration issue and enhance projection accuracy. By consolidating prediction results of multiple captures at different viewpoints, the classification of firefighting assets becomes more robust and accurate compared to single-shot image segmentation. Overall, the proposed approach has been validated through real-world case studies and demonstrated high levels of accuracy in both recognition (average F1-score of 80.5%) and localization (average distance error of 0.524 m). More importantly, the proposed approach offers a scalable solution that can be applied to other types of indoor assets, facilitating more comprehensive enrichment of BIM models.

**5.3 Limitations and future research**

In this research, the limitations are identified from three aspects, including: (1) firefighting asset labeling in the development of the Fire-ART dataset, (2) asset detection and semantic segmentation during the experiments, and (3) alignment between point cloud and BIM as well as BIM enrichment. These limitations deserve further investigation in future research.

First, the labeling work for piping system assets is challenging. Their forms are various, sometimes intertwined, and easily affected by obstacles and joints. In the current dataset, these elements are annotated at a semantic level as a general "piping system," which may not accurately reflect the instance-level pipes. Further research should standardize the labeling rules and enhance the level of semantic details, such as dividing the piping systems into pipe segments and pipe fittings. Another issue in dataset creation is class imbalance. Although the dataset has incorporated diverse sources of images and intentionally capture and supplement the underrepresented asset types, the class imbalance issue still exists, largely due to the natural distribution of these assets in real-world environments. This issue inevitably affects the recognition performance, as reflected in the test set result, where certain firefighting assets (e.g., piping system and firefighting lift switch) gain quite low accuracy, limiting their applicability in practice. Future work should explore advanced data augmentation and synthetic data generation to remedy the scarcity of particular asset types.

Second, there are several unexpected errors in the asset detection and semantic segmentation tasks. In case study one, fire extinguishers with varied functions often appeared in pairs. Although instance-level recognition was achieved during image segmentation, the subsequent camera projection step only transferred semantic labels to the point cloud. As a result, the

DBSCAN algorithm, which relies solely on geometric features, was unable to distinguish between the two closely paired extinguishers, leading to inaccuracies in instance counts. Moreover, some false positive detections indicate that the dataset still needs to be further enriched regarding the fire related objects with similar appearance. Some confusing object pairs discovered in this research are: (a) fire exit signs and other functional signs, and (b) smoke detectors and other indoor sensors like motion detection sensors and air quality sensors. In addition, the accuracy of recognition and segmentation varies across asset categories. In particular, smoke detectors and emergency door releases exhibit lower recall rates due to insufficient training materials, frequent occlusions caused by ambient objects, and challenging indoor lighting conditions. To overcome these issues, future research should conduct instance-level camera projection and enhance the diversity of the dataset.

The following limitations lie in the BIM enrichment stage. The BIM alignment and enrichment process remains partially manual, particularly in the registration of the photogrammetric point cloud and the BIM-derived point cloud. Automating this process is essential for improving workflow efficiency. Furthermore, integrating predefined firefighting asset families into BIM software to automate the model enrichment process will also be considered in future research. Another future work is to expand the scope of semantic enrichment. The proposed approach currently recognizes asset types and calculates their spatial locations, while many semantic details remain underexplored. Take the exit sign as an example, there are various types of exit signs, such as emergency exit sign and the emergency guidance sign. All of them are detected as single exit sign asset at this stage. However, their specific functions (i.e., the directional guidance for evacuation) should be further considered during the BIM enrichment process to provide more valuable information for fire safety management. Moreover, the textual information attached or marked around the firefighting equipment is important, especially for those on the equipment sign and the hidden fire equipment box. Correspondingly, future work should integrate the current approach with text recognition techniques such as OCR to push the boundary of semantic enrichment. Besides, integrating text recognition results could help disambiguate similar-looking objects, thereby reducing the likelihood of false detections.

## 6. Conclusions

Firefighting asset information plays a critical role in building safety management. To support efficient asset management, this research introduces the Fire-ART dataset and proposes a 3D reconstruction approach for semantic enrichment of indoor firefighting assets within BIM models. The Fire-ART dataset contains 15 fundamental asset categories, encompassing 2,626 images and 6,627 instances. Specifically, it is featured by its broad coverage of asset types, extensive image data and annotations, diverse indoor scenes across various spaces and geographical locations, and public accessibility. The release of the Fire-ART dataset provides a valuable resource for model training and benchmarking, thereby contributing to firefighting asset recognition research and applications.

Additionally, this research develops a reconstruction-based approach that converts panoramic photographs into accurate, up-to-date firefighting asset locations within BIM models. By integrating DL-based instance segmentation, modified cube-map conversion, and spherical camera projection, the approach presents a highly automated and cost-effective solution, filling the research gaps in digital documentation of indoor firefighting assets. More importantly, the proposed approach can potentially be extended to other indoor building components, enabling broader applications in facility management and building operations. The approach was validated via two case studies in the UK and Macau SAR, where the result shows promising performance in both firefighting asset recognition and localization, with average F1 scores of 73% and 88% and average distance errors of 0.620 m and 0.428 m, respectively.

Driven by the current limitations, future research should focus on: (1) the continuous development of the building firefighting asset dataset, including a greater number and diversity of asset categories, as well as the incorporation of negative samples to reduce false detections of visually similar objects; (2) the integration with text recognition techniques such as OCR and visual language models to enhance the semantic details and completeness of BIM enrichment outcomes; and (3) the improvement of automated alignment and modeling between the point cloud data and the BIM models based on spatial pattern similarity matching and domain-specific asset placement ruling.

## Acknowledgements

This work was supported by Horizon Europe UKRI Underwrite Innovate, under the grants G115919 – BuildSpace, for Ya Wen, and Ioannis Brilakis from University of Cambridge. The work was also supported by The Science and Technology Development Fund, Macao S.A.R (FDCT project no. 0034/2024/RIB1) for Mun On Wong, Chi Chiu Lam, and Yutong Qiao.

## Data availability

The Fire-ART dataset is available to download from https://1drv.ms/f/c/b84a752d3db5b537/Eg2gXVS-aBpKl2vEMfZ7bxoBC4IwwsYjNxKSubcsQlXjnA?e=0iyNrd.